\documentclass{article}

\usepackage{microtype}
\usepackage{graphicx}
\usepackage{booktabs} 

\usepackage{hyperref}

\usepackage[accepted]{icml2023}

\usepackage{amsmath}
\usepackage{amssymb}
\usepackage{mathtools}
\usepackage{amsthm}

\usepackage[capitalize,noabbrev]{cleveref}

\hypersetup{colorlinks,citecolor={teal}}
\usepackage{multirow}
\usepackage{pifont}
\usepackage{float}
\usepackage{wrapfig}
\newcommand{\cmark}{\ding{51}}
\newcommand{\xmark}{\ding{55}}
\usepackage{makecell}
\usepackage{verbatim}

\usepackage{caption}
\usepackage{subcaption}
\usepackage{arydshln}
\usepackage{bbm}

\usepackage{kotex}
\theoremstyle{plain}

\theoremstyle{definition}

\theoremstyle{remark}

\usepackage[textsize=tiny]{todonotes}

\begin{document}

\twocolumn[
\icmltitle{Fighting Fire with Fire: Contrastive Debiasing without Bias-free Data via Generative Bias-transformation}



\icmlsetsymbol{equal}{*}

\begin{icmlauthorlist}
\icmlauthor{Yeonsung Jung}{kaist}
\icmlauthor{Hajin Shim}{kaist}
\icmlauthor{June Yong Yang}{kaist}
\icmlauthor{Eunho Yang}{kaist,aitrics}
\end{icmlauthorlist}

\icmlaffiliation{kaist}{Graduate School of AI, Korea Advanced Institute of Science and Technology (KAIST), Seoul, South Korea}
\icmlaffiliation{aitrics}{AITRICS, Seoul, South Korea}
\icmlcorrespondingauthor{Yeonsung Jung}{ys.jung@kaist.ac.kr}
\icmlcorrespondingauthor{Eunho Yang}{eunhoy@kaist.ac.kr}

\icmlkeywords{Machine Learning, ICML}

\vskip 0.3in
]



\printAffiliationsAndNotice{}  

\begin{abstract}
Deep neural networks~(DNNs), despite their impressive ability to generalize over-capacity networks, often rely heavily on malignant bias as shortcuts instead of task-related information for discriminative tasks. To address this problem, recent studies utilize auxiliary information related to the bias, which is rarely obtainable in practice, or sift through a handful of bias-free samples for debiasing. However, the success of these methods is not always guaranteed due to the unfulfilled presumptions. In this paper, we propose a novel method, Contrastive Debiasing via Generative Bias-transformation (CDvG), which works without explicit bias labels or bias-free samples. Motivated by our observation that not only discriminative models but also image translation models tend to focus on the malignant bias, CDvG employs an image translation model to transform one bias mode into another while preserving the task-relevant information. Additionally, the bias-transformed views are set against each other through contrastive learning to learn bias-invariant representations. Our method demonstrates superior performance compared to prior approaches, especially when bias-free samples are scarce or absent. Furthermore, CDvG can be integrated with the methods that focus on bias-free samples in a plug-and-play manner for additional enhancements, as demonstrated by diverse experimental results. 
\end{abstract}

\section{Introduction}
Recent advances in deep learning have showcased that DNNs are capable of reaching state-of-the-art performance in various fields of machine learning, such as computer vision~\citep{resnet}, natural language processing~\citep{gpt3}, reinforcement learning~\citep{a3c} and more. However, it is also known that the over-parameterized nature of DNNs not only exposes them to general overfitting but also renders them susceptible to biases present in collected datasets~\citep{torralba11bias}, which are detrimental to the generalizability. 
In supervised learning, neural networks tend to prefer shortcut solutions based on biases rather than real signals~\citep{biasisharmful1, representationbias}. These spurious correlations do not provide task-related information, and DNNs that use these malignant biases, which are easier to perceive compared to the signal, will ultimately fail on future data. For instance, a classifier trained to identify \emph{car racing} images using a dataset dominated by the track will exploit the track road information. However, the classifier will fail to perform well on images of off-road rallies. To this end, debiasing is imperative in utilizing DNNs for real-world applications. 

A tautological solution to the bias problem is to construct a bias-free dataset from the start. 
However, curating a dataset devoid of all bias is extremely costly at best and generally infeasible. A more practical attempt at neutralizing dataset bias is to fortify a dataset with explicit supervision with regard to the bias~\citep{kim2019learningnot, GroupDPO}. However, additional expenditure of human labor in procuring such information cannot be avoided, which renders the option less appealing.

In most cases where such explicit supervision for bias is absent, the following two lines of work are recently proposed. One line of work mitigates the influence of bias by leveraging the bias type~(e.g. texture) ~\citep{rebias, bias_texture, Hex, youngyu} to design bias-oriented auxiliary models or to augment texture-perturbed samples. However, such prior knowledge of bias is by no means guaranteed, and even with such information, designing bias-oriented architectures is not always straightforward. The other line of work focuses on bias-free samples by leveraging the knowledge that malignant biases are learned faster than task-relevant features~\citep{lff, choo, kim2021biaswap}.
However, these methods tend to break down in regimes where bias-free samples are scarce or absent (See Section \ref{subsec:withoutbiasfree}). 

\begin{figure*}[t]
\centering
    \begin{subfigure}[b]{0.48\textwidth}
      \includegraphics[width=\textwidth, height=3cm]{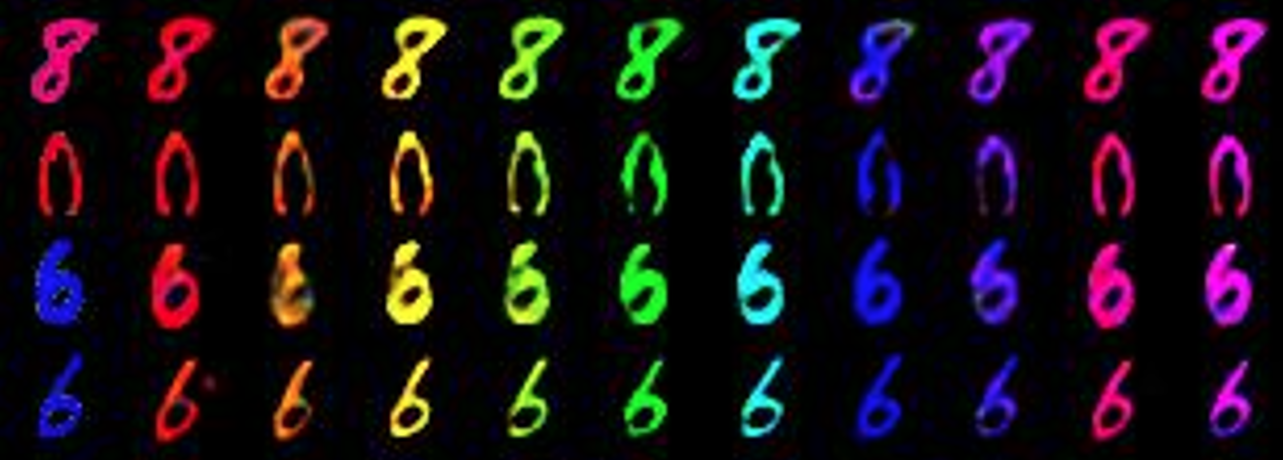}
      \caption{Colored MNIST}
      \label{fig:1}
    \end{subfigure}
    \hfil
    \begin{subfigure}[b]{0.48\textwidth}
      \includegraphics[width=\textwidth, height=3cm]{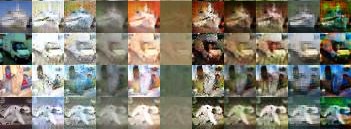}
      \caption{Corrupted CIFAR-10}
      \label{fig:2}
    \end{subfigure}
    \hfil
    \begin{subfigure}[b]{0.48\textwidth}
      \includegraphics[width=\textwidth, height=4cm]{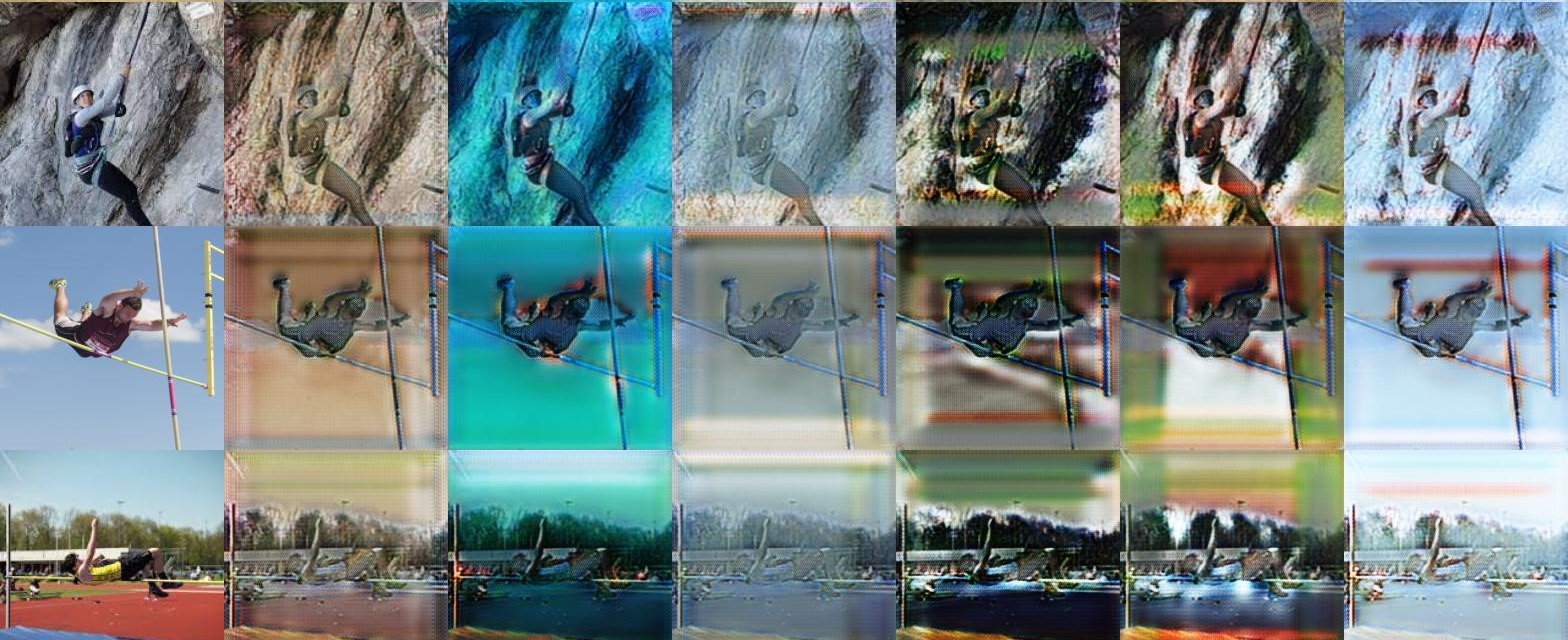}
      \caption{BAR}
      \label{fig:3}
    \end{subfigure}
    \hfil
    \begin{subfigure}[b]{0.48\textwidth}
      \includegraphics[width=\textwidth, height=4cm]{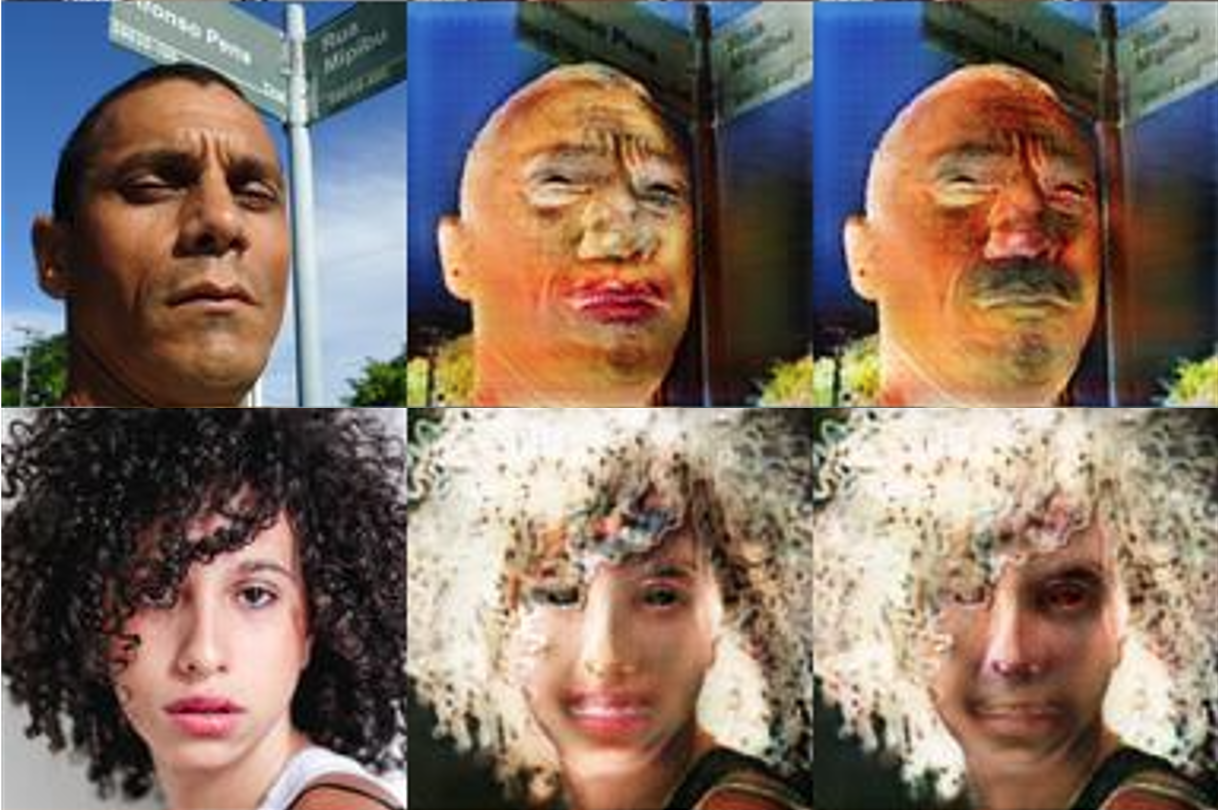}
      \caption{BFFHQ}
      \label{fig:4}
    \end{subfigure}
\caption{Bias-transformed images 
on synthetic and real-world biased datasets. The leftmost column 
contains the original images and each of remaining columns is the bias-transformed image of each target domain. The resulting images show the translated bias attributes such as color (Colored MNIST), texture (Corrupted CIFAR-10), background (BAR), and gender (BFFHQ), respectively. (For details about the datasets, we refer the reader to Appendix \ref{appendix:datasets}.)}
\label{fig:bias-transformed images}   
\vspace{-0.1in}
\end{figure*}

To tackle these shortcomings, we propose a general debiasing method, Contrastive Debiasing via Generative Bias-transformation (CDvG), that can operate even in regimes devoid of bias-free data. CDvG contrasts the bias modes within the dataset against each other to attenuate the influence of bias while effectively learning the task-relevant information contained in all samples. 

Through preliminary experiments, we find that image-to-image translation models based on Generative Adversarial Networks (GANs) favor learning malignant biases over task-relevant signals, as discriminative models are known to do so~(See Section \ref{subsec:biased_gan}).
Motivated by the observation, we train a biased image translation model without bias labels that learns the bias distribution over the signal, allowing us to transform the bias feature of a given input to another bias. 
Using the trained bias-translation model, we synthesize novel views with altered bias features. 

Fighting fire with fire, we pit one bias type against another via contrastive learning. By maximizing agreement between the views with different biases, the model is encouraged to learn bias-invariant representations. Unlike existing methods, CDvG does not require explicit supervision, domain knowledge, or other meta-knowledge, such as the existence of bias-free samples.

Our contributions are three-fold:
\begin{itemize}
    \item We experimentally observe that image-to-image translation models that utilize a domain classifier also favor malignant biases as discriminative models do (Sec. \ref{Sec:Bias}).
    
    \item We propose a novel approach for addressing the highly biased issue without the need for bias labels, bias type information, or even the existence of bias-free samples by employing an image translation model and contrastive learning (Sec. \ref{Sec:method}). 
    
    \item Our method can be integrated with the methods that focus on bias-free samples in a plug-and-play manner for further enhancement. Extensive experiments on diverse datasets empirically demonstrate the effectiveness of our method, especially when bias-free samples are extremely scarce or absent (Sec. \ref{Sec:exp}). 
\end{itemize}

\begin{figure*}[t]
  \centering
  \begin{minipage}[b]{0.58\linewidth}
    \includegraphics[width=\linewidth]{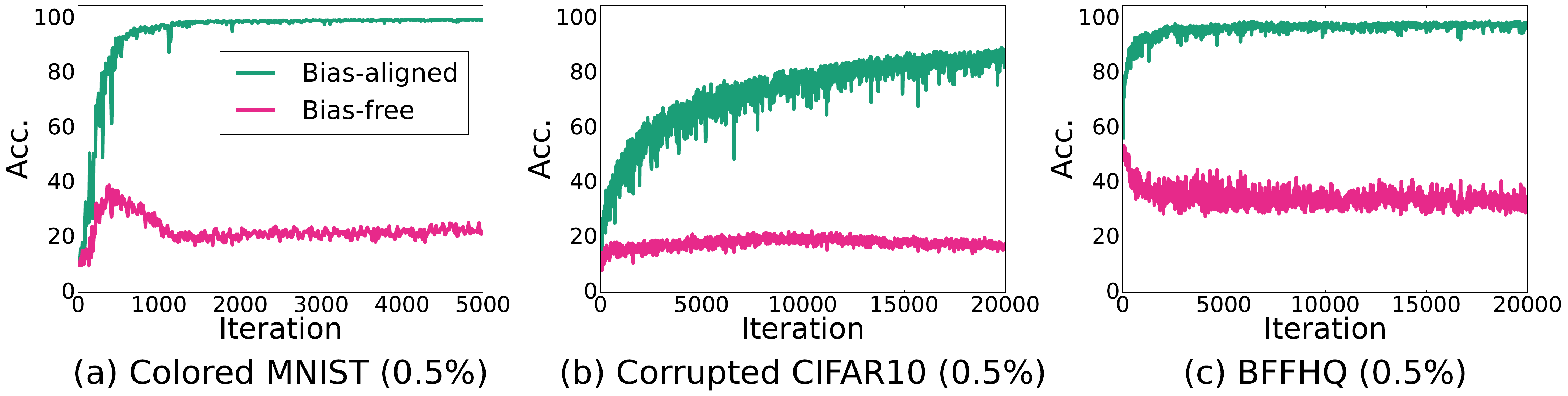}
    \caption{\small Unbiased accuracy during StarGAN training.}
    \label{fig:gan_unbiased}
  \end{minipage}
  \begin{minipage}[b]{0.41\linewidth}
    \includegraphics[width=\linewidth]{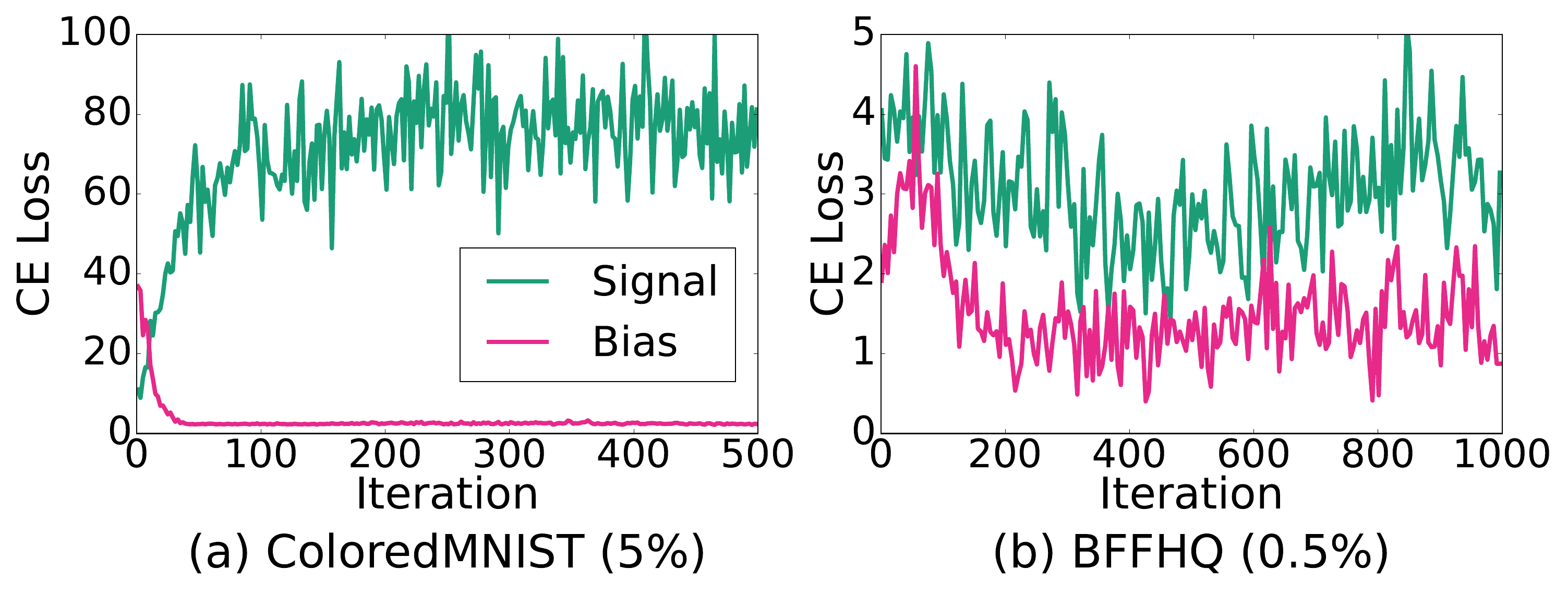}
    \caption{\small CE loss during StarGAN training.}
    \label{fig:gan_eval}
  \end{minipage}
  \vspace{-0.1in}
\end{figure*}

\section{Image Translation Models for Generating Bias-Free Samples}
\label{Sec:Bias}

Previous approaches to addressing the bias problem are to either deploy manpower or algorithms to salvage a sufficient amount of bias-free samples from the contaminated dataset. However, under circumstances where bias-free samples are almost nonexistent, there are not enough bias-free samples to begin with. Instead of salvaging, we first opt to synthesize bias-free samples using image translation models. In other words, we aim to translate samples aligned with the bias to bias-free samples by changing a given bias factor to another.
In this section, we first define our setup in Section \ref{subsec:formulation} and describe our observations of how translation models behave on biased datasets, which is a key part of our algorithm, in Section \ref{subsec:biased_gan}.

\subsection{Setup}
\label{subsec:formulation}

To formally define our target task, we introduce the following random variables: input image $X$, target label $Y$, signal $S$, bias $B$ and other attributes $O$ comprising input $X = (S, B, O)$. Here, $B$ is a feature unrelated to $Y$, that is, $Y$ and $B$ are independent given $S$. We further define the random variable for bias label $Y_B$ which is hidden.

Ideally, the image classification model parameterized by $\theta$ predicts the label based on the signal as $P_\theta(Y|X) = P_\theta(Y|B, S) = P_\theta(Y|S)$. However, when the training data consists of highly but spuriously correlated bias and target, i.e., $P_\theta(Y^{train}|B^{train}) \approx 1$, predicting $Y$ based on $B$ is also one of the possible solutions that can be deemed effective in the training phase.  In this paper, when the sample $x = (s, b, o)$ consists of a correlated signal and a bias (i.e. high $P_\theta(b|s)$), $x$ is called \emph{bias-aligned} and the opposite case is called \emph{bias-free}. We tackle the case where $B$ is easier to perceive than $S$ so that the spurious correlation between $B$ and $Y$ is \emph{malignant} in that the model preferentially takes $B$ as a clue to predict $Y$ over $S$ \citep{lff}. This is obviously an unintended consequence and impairs generalizability due to the discrepancy of the bias-target correlation $Y|B$ between the training and test phases. 

\subsection{Behavior of Image Translation Models Under Bias}
\label{subsec:biased_gan}

We first examine whether GAN-based image translation models~\citep{pix2pix, cyclegan, stargan, starganv2}, which render an image $x$ from a source domain $y$ to a target domain $y'$, are capable of generating bias-free samples. Ideally, they find out the representative characteristics of the target domain $y'$ and combine them with the input image.

Given that discriminative models have been shown to be susceptible to bias, it is not far-fetched to first enquire whether generative models also carry the same frailty. Specifically, it is intuitive to suspect that the family of GANs may be prone to bias due to the presence of a discriminator component, which is known to be bias-pregnable. 

In addition, \citet{cyclegan}, a representative milestone of image translation, presented a number of typical failure modes that when the source domain is an apple and the target domain is an orange, a transformed image is not the orange counterpart of the input, but an apple with the color and texture of an orange.
This implies that the image translation model perceives color and texture rather than shape as the representative traits for the target domain, even without using a highly biased dataset. With this in consideration, it is plausible to speculate that this phenomenon would be exacerbated when handling highly biased datasets.

To verify the impact of bias on image translation models, we examine the behavior of StarGAN~\citep{stargan} on biased datasets. We observe that the translation model interacting with the biased domain discriminator is also prone to translate biases rather than task-related domain features to satisfy the domain classifier (See Figure \ref{fig:bias-transformed images}). The selection of StarGAN is driven by its cost-effectiveness relative to recent models and its status as a well-established, extensively studied in the field. StarGAN's overall size is comparable to that of auxiliary models, such as ResNet18~\citep{resnet}, commonly used in recent debiasing approaches. This makes it a cost-effective alternative to existing baseline models. Furthermore, we found that recent image translation models, which use a multi-task discriminator, are also inherently prone to malignant bias when using a domain classifier. In Appendix~\ref{appendix:starganv2}, we empirically show that replacing the multi-task discriminator with a domain classifier in StarGANv2~\citep{starganv2} leads to a pronounced focus on malignant bias, suggesting that this finding may extend to other recent models.

\begin{figure*}[t]
 \centering
 \includegraphics[width=1.0\textwidth]{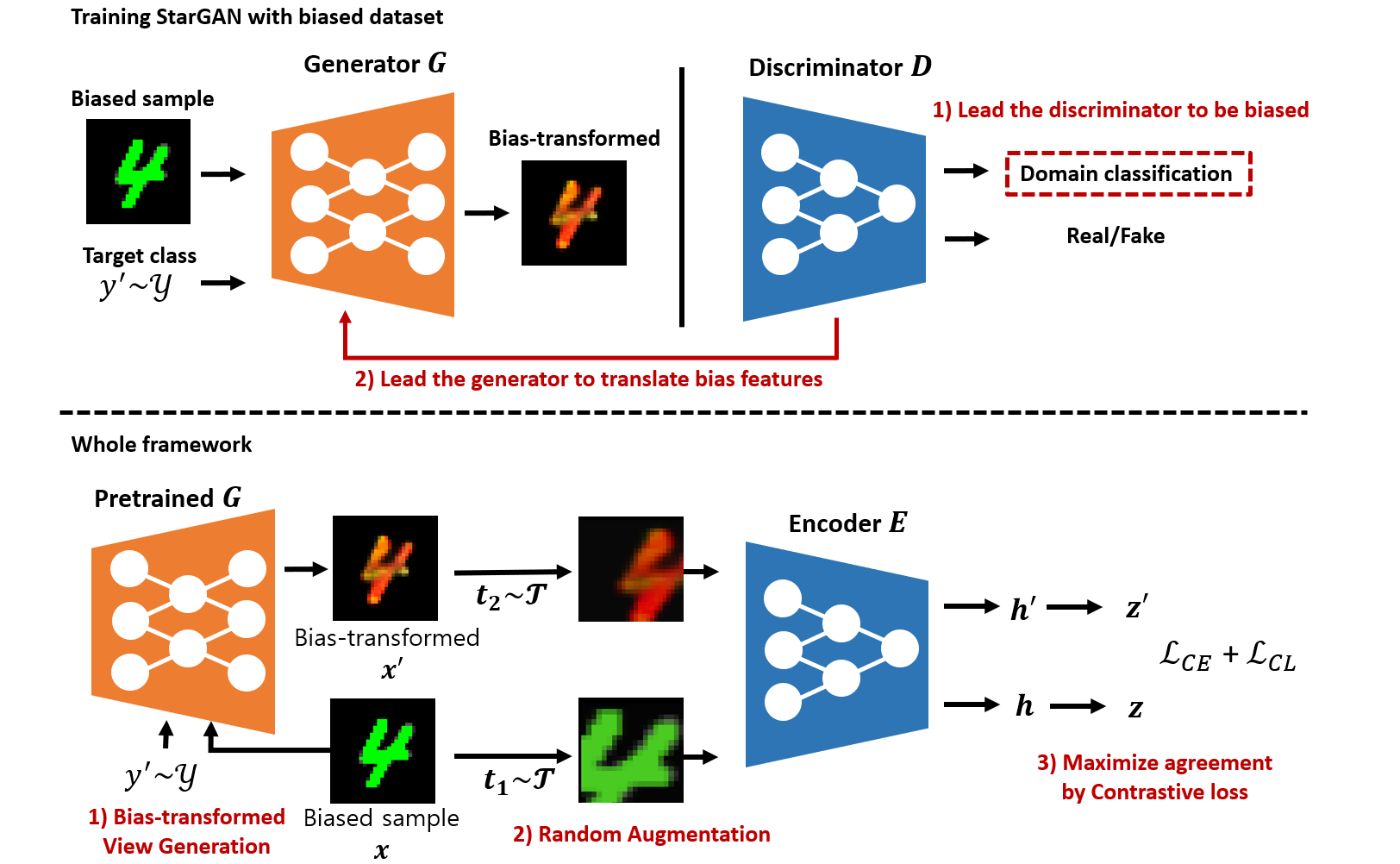}
 \caption{\small Illustration of our Contrastive Debiasing via Generative Bias-transformation~(CDvG).}
\label{fig:main}
\vspace{-0.1in}
\end{figure*}

\paragraph{Discriminators of translation models are also prone to be biased.}
StarGAN introduces an auxiliary domain classifier $D_{cls}$ on top of discriminator $D$ to enable translation between multiple domains. 
The domain classifier $D_{cls}$, trained on the real image with domain labels, learns to classify images with representative traits of the domains by optimizing the domain classification loss of real images $\mathcal{L}_{cls}^{r} = \mathbbm{E}_{(x, y) \sim\mathcal{D}}[-\log D_{cls}(y|x)]$. 

However, on the biased dataset, we observed that $D_{cls}$ had absorbed bias attributes as representative traits of the domains during a training phase. To quantitatively evaluate whether $D_{cls}$ is biased, we measure the classification loss on the unbiased dataset with $D_{cls}$ while training StarGAN on the biased dataset Colored MNIST, Corrupted CIFAR10, and BFFHQ which have color bias, texture bias, and gender bias, respectively (See Figure \ref{fig:bias-transformed images}).
Figure \ref{fig:gan_unbiased} shows the classification losses of bias-aligned and bias-free samples. We observed that the accuracy of bias-aligned samples increases to nearly 100\%, however, the accuracy of bias-free samples is low depending on how malignant the bias of each data set is. Therefore, we conclude that $D_{cls}$ utilizes the bias rather than the task-related features, as discriminative models are known to do so. Please notice that, given the similar architecture, identical biased trainset, and shared classification objective of the translation model's discriminator and the main task's discriminative model, it is natural to expect that both models would learn the same bias. 

\paragraph{A biased domain classifier induces a biased translation generator.}
By optimizing the domain classification loss $\mathcal{L}_{cls}^{f} = \mathbbm{E}_{(x, y) \sim\mathcal{D}, y'\sim\mathcal{Y}}[-\log D_{cls}(y'|G(x, y'))]$ of translated fake images $G(x, y')$, biased $D_{cls}$ induces the generator 
$G$ to translate the bias attribute of image $x$ into the other bias correlated to the randomly sampled target domain $y'$ rather than the task-relevant signal. As a result, $G$ becomes a bias-translator. 
This phenomenon becomes more noticeable as the bias is malignant for $D_{cls}$ - that is, the more scarce the bias-free samples are or the easier the bias is to perceive. To quantitatively evaluate whether the translation model truly favors learning biases over task-relevant signals, we measure the classification loss of translated images $x' = G(x, y')$ with the bias classifier $C_B$ and the signal classifier $C_S$ while training StarGAN on the biased dataset Colored MNIST and BFFHQ. As alternatives of oracles, the classifiers $C_B$ and $C_S$ are trained with the bias label $y_B$ and the true class label $y$, respectively, on the unbiased Colored MNIST and FFHQ. 
The bias and signal loss are defined as $\mathbbm{E}_{(x,y)\sim\mathcal{D}, y'\sim Cat (|\mathcal{Y}|)}[\mathcal{L}_{CE}(C_{B}(x'), y')]$ and $\mathbbm{E}_{(x,y)\sim\mathcal{D}, y'\sim Cat(|\mathcal{Y}|)}[\mathcal{L}_{CE}(C_{S}(x'), y')]$, respectively.

As expected, the results depicted in Figure~\ref{fig:gan_eval} demonstrate that the bias is more favorable to the translation model. For Colored MNIST, the bias loss quickly declines to zero, but the signal loss increases rapidly since $G$ concentrates on the color biases rather than digits. In the case of BFFHQ, the bias is less malignant, resulting in a relatively small gap between signal and bias loss, but still places significant emphasis on the bias. 

We also present the qualitative results in Figure~\ref{fig:bias-transformed images}. 
Interestingly, we found that the translated results retain their contents to some extent and color, texture, background (e.g. rock wall for climbing, water for fishing), 
and gender characteristics (e.g. makeup for female, beard for male), which are the respective bias features of the datasets, are altered. 

Although the generated images of BFFHQ in Figure \ref{fig:4} appear dissimilar to real images when viewed by humans, the core contents remain discernible while the gender-specific features emerge as pink lips and eye makeup in the second column and thick eyebrows and beard in the third column. Furthermore, it is notable that the biased classifier judges that the bias-transformed images are well transformed into the target label with a high degree of confidence. These observations verify that there is a chance that image translation models trained on biased datasets can be employed to synthesize bias-free samples for debiasing.

\section{Method}
\label{Sec:method}
In this section, we empirically show that the recent debiasing methods break down when the assumption of the availability of bias-free samples is not met in Section \ref{subsec:withoutbiasfree}. Acknowledging this limitation, we propose a novel debiasing method, referred to as CDvG, in Section \ref{subsec:generative bias-transformation} leveraging our empirical findings (Section \ref{subsec:biased_gan}) that image translation models are prone to be biased.

\subsection{Learning without Bias-free Samples} \label{subsec:withoutbiasfree}
Although recent debiasing methods~\citep{rebias, youngyu, lff, choo, kim2021biaswap} work as intended when they can exploit the bias-free samples, they neglect the case where such bias-free samples are extremely scarce or absent. Thus, in regimes where a presumption that enough bias-free samples exist does not hold, their behaviors are practically unknown and are likely to break down. More specifically, in Table~\ref{table:main_synthetic}, we empirically demonstrate that previous methods fail to effectively debias the model~(showing low performance, almost comparable to Vanilla) when the proportion of bias-free samples is 0\%.

Furthermore, we investigate the sample reweighting scheme~\citep{lff, choo} that aims to balance bias-free and bias-aligned samples. We observed that as the proportion of bias-free samples decreases, the debiasing effect also decreases, due to the inadequate emphasis placed on bias-free samples. The detailed descriptions and results are deferred to Section \ref{sec:supp_withoutbiasfreesamples}. Based on these observations, we consider a more general task of learning-under-bias that handles the situations where there are no to only a few bias-free samples as well as the cases where such samples are sufficiently provided.

\renewcommand{\arraystretch}{1.1}
\begin{table*}[t]
\caption{The average and the standard deviation of accuracy over 3 runs. Ratio(\%) represents that the proportion of bias-free samples.
{\cmark} indicates that the model exploits auxiliary information of the bias and {\xmark} does not.
}
\begin{center}
\setlength{\tabcolsep}{2.5pt} 
\resizebox{\linewidth}{!}{
\begin{tabular}{cccccccccc}
\toprule
\multirow{2}{*}{Dataset} & \multirow{2}{*}{Ratio(\%)} & Vanilla & ReBias & LfF & DisEnt & BiaSwap & BPA & CDvG & CDvG+LfF \\
\cline{3-10}
&& \xmark & \cmark & \xmark & \xmark & \xmark & \xmark & \xmark & \xmark\\
\cline{1-10}
\rule{0pt}{2ex}

\multirow{5}{*}{\thead{Colored\\MNIST}} 
& 0.0 & 12.53\tiny{$\pm 0.92$} & 14.64\tiny{$\pm 0.50 $} & 13.16\tiny{$\pm 1.87$} & 11.65\tiny{$\pm 0.61$} & - & 10.08\tiny{$\pm 0.04$} & \underline{96.16}\tiny{$\pm 0.48$} & \textbf{96.48}\tiny{$\pm 0.19$}\\

& 0.5 & 39.12\tiny{$\pm 0.91 $} & 70.47\tiny{$\pm 1.84$} & 66.54\tiny{$\pm 3.80 $} & 62.13\tiny{$\pm 3.96$} & 85.76 & 54.52\tiny{$\pm 3.39$} & \underline{95.73}\tiny{$\pm 0.14$} & \textbf{96.20\tiny{$\pm 0.12$}}\\
& 1.0 & 56.02\tiny{$\pm 1.81 $} & 87.40\tiny{$\pm 0.78$} & 79.83\tiny{$\pm 2.23 $} & 75.49\tiny{$\pm 0.21 $} & 83.74 & 72.63\tiny{$\pm 0.27$} & \underline{96.13}\tiny{$\pm 0.48$} & \textbf{96.45\tiny{$\pm 0.19$}} \\
& 2.0 & 69.32\tiny{$\pm 0.22 $} & 92.91\tiny{$\pm 0.15$} & 82.66\tiny{$\pm 0.39 $} & 80.08\tiny{$\pm 0.45 $} & 85.29 & 78.52\tiny{$\pm 0.59$} & \underline{96.90}\tiny{$\pm 0.22$} & \textbf{96.97\tiny{$\pm 0.17$}}\\
& 5.0 & 83.93\tiny{$\pm 0.89$} & \underline{96.96}\tiny{$\pm 0.04$} & 83.30\tiny{$\pm 1.23 $} & 85.00\tiny{$\pm 1.17$} & 90.85 & 85.30\tiny{$\pm 0.93$} & 96.73\tiny{$\pm 0.05$} & \textbf{96.98}\tiny{$\pm 0.05$}\\
\cline{1-10}
\rule{0pt}{2ex}

\multirow{5}{*}{\thead{Corrupted\\CIFAR-10}} 
& 0.0 & 16.05\tiny{$\pm 0.13$} & 21.93\tiny{$\pm 0.37$} & 15.88\tiny{$\pm 0.45$} & 18.76\tiny{$\pm 0.88$} & - & 17.14\tiny{$\pm 1.54$} & \underline{28.44}\tiny{$\pm 0.21$} & \textbf{29.24\tiny{$\pm 0.47$}}\\
& 0.5 & 20.87\tiny{$\pm 0.34$} & 22.27\tiny{$\pm 0.41$} & 25.58\tiny{$\pm 0.23$} & 28.62\tiny{$\pm 1.74$} & 29.11 & 25.50\tiny{$\pm 1.03$} & \underline{31.50}\tiny{$\pm 0.33$} & \textbf{39.07\tiny{$\pm 0.27$}}\\
& 1.0 & 24.05\tiny{$\pm 0.61$} & 25.72\tiny{$\pm 0.20$} & 30.68\tiny{$\pm 0.50$} & 32.31\tiny{$\pm 0.03$} & 32.54 & 26.86\tiny{$\pm 0.69$} & \underline{33.25}\tiny{$\pm 0.20$} & \textbf{43.81\tiny{$\pm 0.35$}}\\
& 2.0 & 29.47\tiny{$\pm 0.20$} & 31.66\tiny{$\pm 0.43$} & \underline{37.96}\tiny{$\pm 1.09$} & 36.51\tiny{$\pm 2.34$} & 35.25 & 27.47\tiny{$\pm 1.46$} & 35.16\tiny{$\pm 0.23$} & \textbf{47.45\tiny{$\pm 0.07$}}\\
& 5.0 & 41.12\tiny{$\pm 0.16$} & 43.43\tiny{$\pm 0.41$} & \underline{48.49}\tiny{$\pm 0.16$} & 46.41\tiny{$\pm 0.62$} & 41.62 & 34.29\tiny{$\pm 2.20$} & 42.75\tiny{$\pm 0.19$} & \textbf{52.31\tiny{$\pm 0.13$}} \\

\cline{1-10}
\rule{0pt}{2ex}

\multirow{2}{*}{BFFHQ} 
& 0.0 & 37.93\tiny{$\pm 0.96$} & 43.47\tiny{$\pm 0.74$} & 39.67\tiny{$\pm 1.00$} & 38.13\tiny{$\pm 2.13$} & - & 48.20\tiny{$\pm 1.40$} & \underline{48.80}\tiny{$\pm 1.18$} & \textbf{49.60}\tiny{$\pm 1.18$} \\
& 0.5 & 52.40\tiny{$\pm 1.88$} & 56.80\tiny{$\pm 1.56$} & \underline{58.07}\tiny{$\pm 0.82$} & 54.33\tiny{$\pm 0.92$} & - & 51.40\tiny{$\pm 2.98$} & 56.80\tiny{$\pm 0.33$} & \textbf{62.20\tiny{$\pm 0.45$}}\\
\bottomrule
\end{tabular}
} 
\end{center}
\label{table:main_synthetic}
\vspace{-0.2in}  
\end{table*}

\subsection{Contrastive Debiasing without Bias-free Data via Bias-transformed Views}
\label{subsec:generative bias-transformation}

Motivated by the findings of Section~\ref{Sec:Bias}, we propose a novel debiasing method referred to as Contrastive Debiasing via Generative Bias-transformation (CDvG). The method employs the biased translation model to transform an image to have different biases corresponding to other classes and integrates it with the contrastive learning framework. The contrastive loss is employed to encourage the learning of bias-invariant representations without the need for explicitly identifying bias-free data. The bias-transformed views are contrasted against each other to attenuate the bias while effectively learning the features that are relevant to the task. The whole process is outlined in the following paragraphs and summarized in Algorithm \ref{alg_1}.

First, we train StarGAN on a given biased training dataset to obtain the bias-transformation generator $G$. 
The bias-transformation generator $G$ is then utilized to generate diverse bias-transformed views $x' = G(x, y')$ of an input image $x$ with a target label $y'$, where $y'$ is uniformly sampled for every iteration (Step 1 of the whole framework in Figure \ref{fig:main}, Line 4 and 5 in Alg. \ref{alg_1}). Bias-translation with $G$ is only applied during training, not in the testing phase, due to the distribution shifts. 

\setlength{\textfloatsep}{0.25in}
\begin{algorithm}[t]
    \caption{Contrastive Debiasing via Generative Bias-transformation (CDvG)}
    \label{alg_1}
    {
    \begin{algorithmic}[1]
        \STATE {\bfseries Input:} Encoder $E$, Projection head $H$, Classifier $C$, Biased generator $G$, Augmentation family $\mathcal{T}$
        \STATE {\bfseries Data:} Training set $\mathcal{D} = \{(x, y)\} \subset \mathcal{X} \times \mathcal{Y} $

        \FOR {\textup{minibatch} $\{(x_k, y_k)\}_{k=1}^N $}
        \STATE $y'_k \sim Categorical(|\mathcal{Y}|)$
        \STATE $x_k' = G(x_k, y'_k)$ 
        \STATE
        \STATE $t_1\sim \mathcal{T}, t_2\sim \mathcal{T}$
        \STATE $\tilde{x}_{k},  \tilde{x}_{k}' = t_1(x_k), t_2(x_k')$

        \STATE
        \STATE {\bfseries Update} $E, C, H$ to minimize
        \STATE { $\sum_k ( \mathcal{L}_{CE}(C(E(\tilde{x}_{k})), y_k) + \mathcal{L}_{CE}(C(E( \tilde{x}_{k}')), y_k)) + \mathcal{L}_{CL}(E, H)$}
        \ENDFOR
    \end{algorithmic}
    }
\end{algorithm}

After the bias-transformation step, we additionally apply the random augmentation operators $t_1$ and $t_2$ to the original and the bias-transformed images respectively as $\tilde{x} = t_1(x)$, $\tilde{x}' = t_2(x')$, where $t_1$ and $t_2$ are sampled from the same augmentation family $\mathcal{T}$ (Step 2 of the whole framework in Figure \ref{fig:main}, Line 7 and 8 in Alg. \ref{alg_1}). According to \citet{simclr}, strong augmentations are essential to the performance of the contrastive learning framework as they prevent the encoder from easily finding a clue about the fact that two views come from the same image. By following \citet{simclr}, $\mathcal{T}$ is composed of the following sequential augmentations: random resized cropping and random horizontal flipping. Please note that we did not adopt color distortion methods because most biases are related to color.

Finally, the resulting learning objective is given by the combination of the cross entropy loss and the contrastive loss (Step 3 of the whole framework in Figure \ref{fig:main}, Line 11 in Alg. \ref{alg_1}). First, the encoder $E$ and the following classifier $C$ are optimized to minimize the cross entropy loss ${L}_{CE}$ which is applied to both $\tilde{x}$ and $\tilde{x}'$:
\begin{align}
\min_{E,C} \mathcal{L}_{CE}(C(E(\tilde{x})), y) + \mathcal{L}_{CE}(C(E(\tilde{x}')), y).\; \nonumber
\end{align}
Also, we train the encoder $E$ and the projection head $H$ to minimize the contrastive loss $\mathcal{L}_{CL}$ to tie the original view $\tilde{x}$ and the bias-transformed view $\tilde{x}'$ as $\mathcal{L}_{CL}(E, H) = \sum_{k=1}^N \ell(2k-1, 2k) + \ell(2k, 2k-1)$. The loss $\ell(i, j)$ for a positive pair $(i, j)$ is defined as
\begin{align}
 \ell(i, j) = -\log\frac{\exp(sim_{i,j}/\tau)}{\sum_{k=1}^{2N}\mathbbm{1}_{k\neq i} \exp(sim_{i,k}/\tau)},\; \nonumber
 \end{align}
Where $sim_{i,j} = z_i^\top z_j / (||z_i|| ||z_j||)$ is the cosine similarity, and $z_i = H(E(\tilde{x}_i))$ is a projected representation of $\tilde{x}_i$ with the base encoder $E$ followed by the projection head $H$. By tying the original view $\tilde{x}$ and the bias-transformed view $\tilde{x}'$ as a positive pair, the encoder attempts to attenuate biases while emphasizing the true signals shared by the views via maximizing the mutual information between their latent representations. 

\renewcommand{\arraystretch}{1.1}
\begin{table*}[t]
\caption{ImageNet-9 dataset.}
\vspace{-0.1in}
\begin{center}
\resizebox{0.65\linewidth}{!}{
\begin{tabular}{clccccc}
\toprule
Dataset & Test type & Vanilla & RUBi & LfF & Rebias & CDvG+LfF \\
\cline{1-7}
\multirow{3}{*}{IN-9}
& Biased & 94.0\tiny{$\pm 0.1$} & 93.9\tiny{$\pm 0.2$} & 91.2\tiny{$\pm 0.1$} & 94.0\tiny{$\pm 0.2$} & \textbf{95.2\tiny{$\pm 0.1$}} \\

& Unbiased & 92.7\tiny{$\pm 0.2$} & 92.5\tiny{$\pm 0.2$} & 89.6\tiny{$\pm 0.3$} & 92.7\tiny{$\pm 0.2$} & \textbf{94.5\tiny{$\pm 0.1$}} \\

& IN-a & 30.5\tiny{$\pm 0.5$} & 31.0\tiny{$\pm 0.2$} & 29.4\tiny{$\pm 0.8$} & 30.5\tiny{$\pm 0.2$} 
 & \textbf{34.6\tiny{$\pm 0.4$}} \\

\bottomrule
\end{tabular}
} 
\end{center}
\label{table:largedata}
\vspace{-0.1in}
\end{table*}

\renewcommand{\arraystretch}{1.1}
\begin{table*}[t]
\caption{Waterbirds dataset.}
\vspace{-0.1in}
\begin{center}
\resizebox{0.8\linewidth}{!}{
\begin{tabular}{cccccc||ccc}
\toprule
\multirow{2}{*}{Dataset} & \multirow{2}{*}{Test type} & \multicolumn{4}{c}{ResNet-18} & \multicolumn{3}{c}{ResNet-50} \\ 

& & Vanilla & LfF & BPA & CDvG+LfF & ERM & EIIL & CDvG+LfF\\

\midrule

\multirow{2}{*}{Waterbirds} & Unbiased & 84.63 & 85.48 & \textbf{87.05} & 86.25 & \textbf{97.30} & 96.90 & 91.30 \\
& Worst-group & 62.39 & 68.02 & 71.39 & \textbf{74.92}& 60.30 & 78.70 &  \textbf{84.80} \\
\bottomrule
\end{tabular}
} 
\end{center}
\label{table:waterbird}
\end{table*}

\section{Experiments}
\label{Sec:exp}
To validate the effectiveness of CDvG compared to recent debiasing methods, we conduct image classification experiments on standard benchmark datasets for debiasing. We first outline the experimental setup, including the datasets and baselines in Section~\ref{settings}. Then, we present the main results, which consist of comprehensive comparisons between our method and the baselines across various datasets in Section~\ref{main_results}, and provide detailed analysis of the learned bias-invariant representations by visualizations in Section~\ref{qualitative_results}. Additionally, we conducted ablation studies to demonstrate the contributions of each component of CDvG toward the performance improvements in Section~\ref{ablation}. Also, detailed information regarding the datasets is provided in Appendix~\ref{appendix:datasets}, while implementation specifics are covered in Appendix~\ref{appendix:implementation}."

\subsection{Experimental settings}\label{settings}
\paragraph{Dataset}
We experiment on \{Colored MNIST, Biased MNIST, Corrupted CIFAR-10\} and \{
BFFHQ, ImageNet-9 (IN-9), Waterbirds\} which are synthetic datasets injected with synthetic biases and real-world datasets with natural biases, respectively. By setting the proportion of bias-free samples to a range of 0-5\%, we evaluate the performance considering the highly biased setting (0.5-5\%) as following the convention and, moreover, the most challenging scenario (0\%) where bias-free samples are absent. The datasets with 0\% bias-free samples are constructed by excluding bias-free samples from the datasets with a ratio of 0.5\%.
\paragraph{Baselines}
To evaluate the performance of CDvG, we conducted comprehensive comparisons with various categories of debiasing methods. Specifically, we compared CDvG against HEX~\citep{Hex} and Rebias~\citep{rebias}, which leverage domain knowledge to address bias, as well as LfF~\citep{lff}, which intentionally trains an auxiliary biased classifier to identify and utilize bias-free samples. Moreover, we assessed DisEnt~\citep{choo} and BiaSwap~\citep{kim2021biaswap}, which build upon the bias-free sample selection scheme employed by LfF, aiming to incorporate other biases into the instances. Regarding BiaSwap, we relied on available performance directly from the corresponding paper while leaving blank spaces for the others  due to the unavailability of the official code. Additionally, we examined BPA~\citep{pseudoattr} and EIIL~\citep{EIIL} for further comparison.


\subsection{Main Results} \label{main_results}
We report the results on the standard benchmark datasets to validate the effectiveness of CDvG for debiasing when bias labels are not available. We observed that the CDvG demonstrated superior performance when bias is more malignant, specifically when bias-free samples are scarce(0.5\%) or absent(0\%). Furthermore, to leverage bias-free samples, we propose to combine CDvG and reweighting-based methods such as LfF~\citep{lff} (‘CDvG+LfF’ in Table 1). The combined version is consistently superior to the baselines with a low standard deviation in Table~\ref{table:main_synthetic}. In more detail, in the training process, we trained CDvG by concurrently applying LfF’s sample-wise weighting to CE loss for both original images and bias-translated images. Therefore, ‘CDvG+LfF’ allows us to utilize both bias-aligned and bias-free samples to debias by translating bias-aligned samples into bias-free samples (CDvG) and giving more weights to bias-free samples (LfF). Please note that CDvG can be easily integrated with other debiasing methods in a plug-and-play manner.

\renewcommand{\arraystretch}{1.3}
\begin{table}[b]
\caption{Biased MNIST dataset.}
\vspace{-0.1in}
\begin{center}
\resizebox{\linewidth}{!}{
\begin{tabular}{ccccccc}
\toprule
{Dataset} & ERM & SD & UpWt & GroupDPO &  PGI & CDvG \\
\cline{1-7}
\rule{0pt}{2ex}
BiasedMNIST & 36.8 & 37.1 & 37.7 & 19.2 & 48.6 & \textbf{49.48} \\
\bottomrule
\end{tabular}
}
\end{center}
\label{table:bmnist}
\vspace{-0.2in}
\end{table}

\paragraph{Synthetic datasets} In Table~\ref{table:main_synthetic}, we observe that the CDvG performs well on Colored MNIST and Corrupted CIFAR-10 when the bias-free samples are scarce or absent, i.e., when the biases are more malignant. CDvG+LfF, which is our integrated method, outperforms the baselines on the overall ratio by a large margin.



\paragraph{Real-world datasets} We conducted experiments on BFFHQ, IN-9, and Waterbirds to evaluate the performance in a real-world setting. CDvG+LfF shows improved performance with small standard deviations on BFFHQ compared to state-of-the-art and also shows improved or comparable performance on IN-9 and Waterbirds in Table~\ref{table:largedata} and Table~\ref{table:waterbird}, respectively. For Waterbirds, we borrow the performance on ResNet-18 and ResNet-50 directly from BPA and EIIL, respectively.

\paragraph{Multiple biases datasets} To verify our method can handle multiple types of biases, we conducted experiments on Biased MNIST which has ‘digit color’, ‘type of background texture’, ‘background color’, ‘co-occurring letters’, and ‘colors of the co-occurring letters’ as biases. We confirmed that CDvG shows better performance than baselines when multiple biases exist in Table~\ref{table:bmnist}. Also, we confirm that the translation model uncovers all sources of biases according to their intensity. We borrow the performance of SD~\citep{sd}, GroupDPO~\citep{GroupDPO}, and PGI~\citep{pgi} on ResNet-18 from \citet{occamnet}.

In summary, our method can handle various types of synthetic and real-world biases and works well on a wide range of bias ratios, especially when bias-free samples are scarce and even absent compared with state-of-the-art methods. 
Overall consistent results empirically demonstrate that our method debias better than previous approaches focusing only on bias-free samples.

\begin{figure}[t]
\centering
\captionsetup{width=1.0\linewidth}
\includegraphics[width=1.0\linewidth]{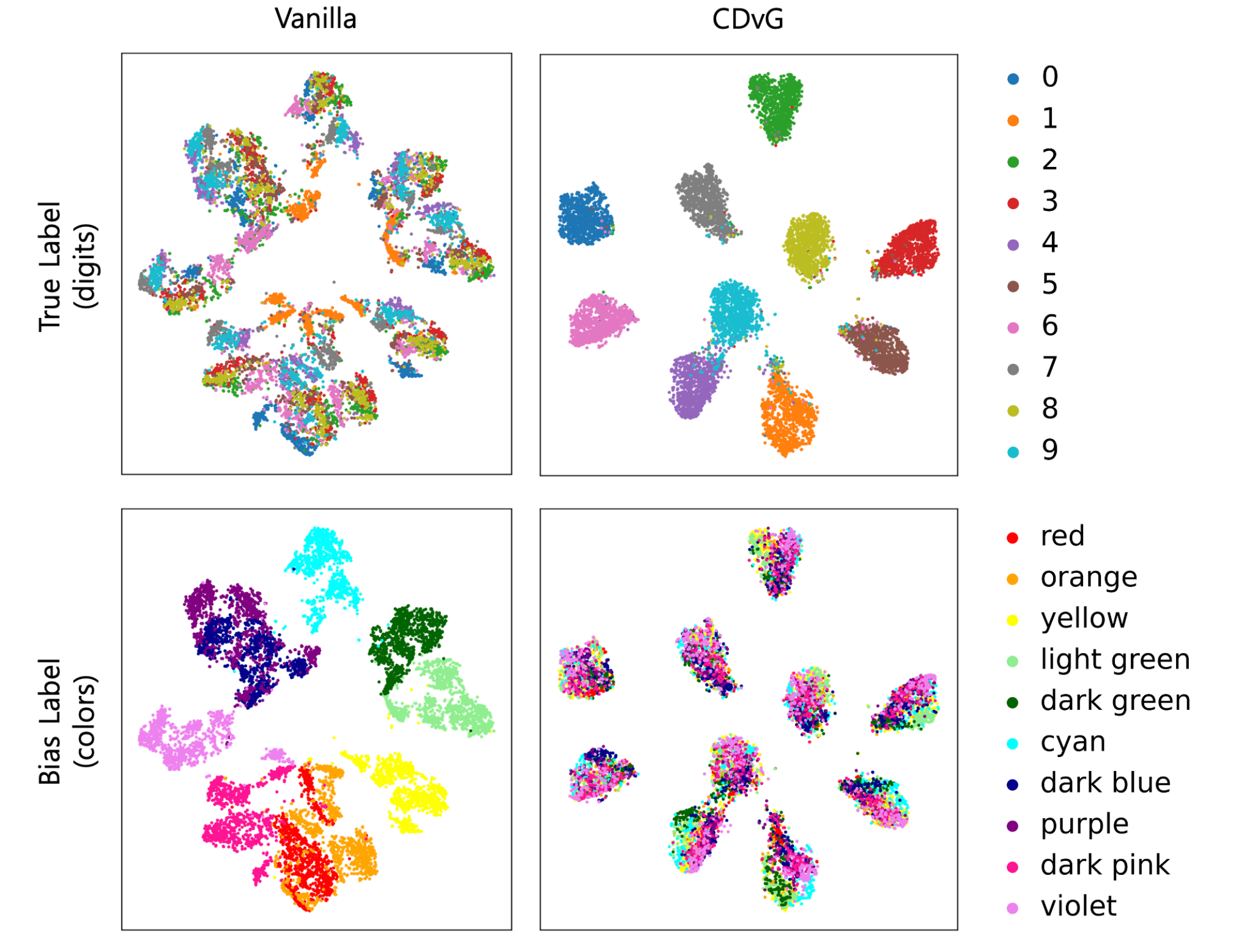}
\caption{Visualization of learned representations of the vanilla model and CDvG using t-SNE in Colored MNIST.}
\label{fig:tsne}
\end{figure}

\subsection{Visualizations} \label{qualitative_results}

We explore the interpretability of the learned bias-invariant representations using t-SNE~\citep{tsne} and GradCAM~\citep{gradcam} on Colored MNIST and BFFHQ datasets, respectively. Our results on t-SNE show that the vanilla model heavily relies on the biased attribute 'color' for classification, as shown in the first column in Figure \ref{fig:tsne}. In contrast, CDvG focuses more on the true label 'digits', as shown in the second column. This indicates that CDvG is better able to learn bias-invariant representations and is less dependent on biased attributes.

\begin{figure}[t]
\centering
\captionsetup{width=1.0\linewidth}
\includegraphics[width=1.0\linewidth]{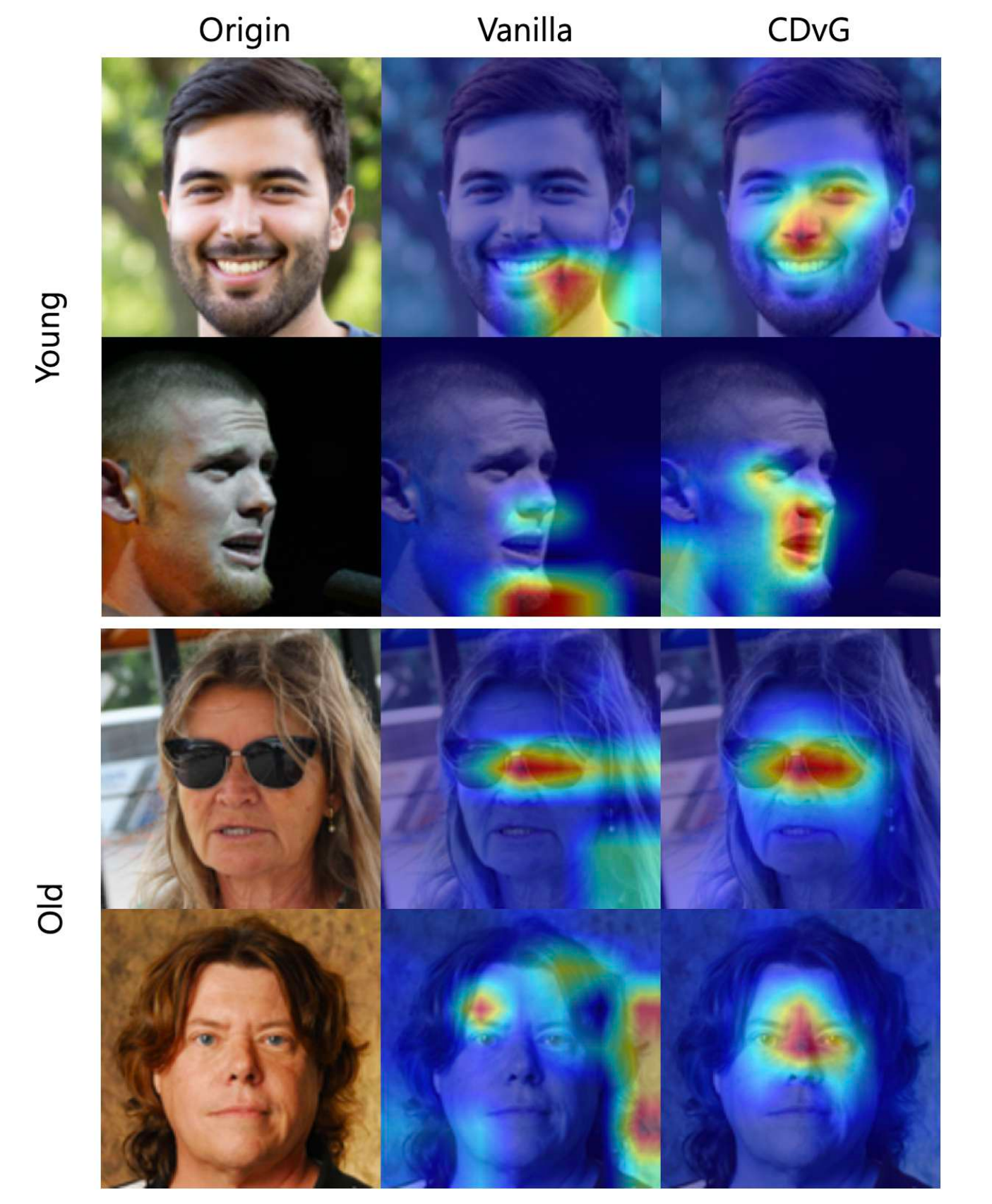}
\caption{Visualization of Grad-CAM heatmaps for the labels 'Young' and 'Old' of BFFHQ using the vanilla model and the proposed CDvG.}
\label{fig:gradcam}
\end{figure}

Similarly, in the case of GradCAM, we found that the vanilla model heavily relies on biased attributes 'gender' such as the beard of a male and the hair of a female, while CDvG is less dependent on these attributes and instead focuses more on facial features in Figure~\ref{fig:gradcam}. These findings demonstrate that CDvG can learn more bias-invariant representations than the vanilla model.

\subsection{Ablation study} \label{ablation}
We study the effectiveness of each component of CDvG framework. As shown in Table~\ref{table:ablation}, CDvG without bias-transformation generator (CDvG w/o $G$), which is adding augmentation operator $\mathcal{T}$ and contrastive loss to vanilla, shows degradation in performance. This is due to the fact that the use of an $\mathcal{T}$ alone, without the bias-transformation generator, results in augmented samples that are still biased and ultimately contribute to the model becoming biased.
However, when only the bias-transform generator $G$ is used for augmentation (CDvG w/o CL), there are 8.21\%, 5.37\%, 4.07\%, 2.22\%, and 1.43\% improvements for each ratio, respectively. This demonstrates that the bias-transformation generator effectively augments bias-transformed images to some extent.

Despite this, it is important to note that the bias-transformation generator $G$ is not always a perfect model, as it may not fully translate biases due to the intensity of the bias. Therefore, it is essential to induce bias-invariance and capture the true signal shared by the bias-transformed view and the original view by contrastive loss. The results of the ablation study show that the use of both the bias-transformation generator and contrastive loss (CDvG) results in the most significant improvement in performance, demonstrating that our whole framework effectively addresses the bias problem rather than a single component. 

\begin{table}[h]
\caption{Ablation study for bias-transformed view generation (w/o G) and for contrastive loss (w/o CL).}
\vspace{-0.1in}
\begin{center}
\large
\resizebox{1.0\columnwidth}{!}{%
\begin{tabular}{cccccc}
\toprule
{Dataset} &  Ratio(\%) & Vanilla & CDvG w/o $G$ & CDvG w/o CL & CDvG \\
\cline{1-6}
\rule{0pt}{2ex}

\multirow{5}{*}{\thead{Corrupted\\CIFAR-10}}
& 0.0 & 16.04\tiny{$\pm 0.13$} & 18.29\tiny{$\pm 0.90$} & 24.25\tiny{$\pm 0.15$} & \textbf{28.44\tiny{$\pm 0.21$}} \\
& 0.5 & 20.87\tiny{$\pm 0.34$} & 20.75\tiny{$\pm 0.21$} & 26.24\tiny{$\pm 0.12$} & \textbf{31.50\tiny{$\pm 0.33$}} \\
& 1.0 & 24.05\tiny{$\pm 0.61$} & 21.37\tiny{$\pm 0.29$} & 28.12\tiny{$\pm 0.21$} & \textbf{33.25\tiny{$\pm 0.20$}} \\
& 2.0 & 29.47\tiny{$\pm 0.20$} & 24.50\tiny{$\pm 0.34$} & 31.69\tiny{$\pm 0.41$} & \textbf{35.16\tiny{$\pm 0.23$}} \\
& 5.0 & 41.12\tiny{$\pm 0.16$} & 31.13\tiny{$\pm 0.56$} & 42.55\tiny{$\pm 0.50$} & \textbf{42.75\tiny{$\pm 0.19$}} \\
 

\bottomrule
\end{tabular}
}
\end{center}
\label{table:ablation}
\vspace{-0.2in}
\end{table}

\section{Related Work}
In real-world datasets across diverse domains, various kinds of incidental biases are strongly but spuriously correlated with task-related information. However, when the bias factors are more noticeable and easier to learn than the task-related signals, DNNs tend to lean on such biases \citep{lff}, causing failures in generalization. To counter this effect, several lines of work were developed.

\paragraph{Explicit bias supervision} can be used to screen or mitigate the influence of bias. \citet{kim2019learningnot} uses bias supervision to train an auxiliary network that helps in reducing the influence of bias. \citet{GroupDPO} uses bias supervision to group data samples for grouped distributionally robust optimization.
\citet{youngyu} propose bias-contrastive loss and bias-balanced regression that encourages the model to pull together the samples in the same class with different bias features with balancing the 
 target-bias distribution.

\paragraph{Domain knowledge of biases.}  When acquiring bias supervision is impractical, we can leverage domain-specific knowledge about bias type. For example, it was shown that ImageNet-trained classifiers exploit texture information in the image rather than information contained in the object of interest~\citep{gatys, brendel2018approximating}. 
Utilizing this fact, \citet{bias_texture} construct an augmented dataset by applying various textures to the 
images for texture debiasing. Also, \citet{Hex} and \citet{rebias} 
exploit an auxiliary model which is carefully designed to capture biases over signals.
However, these methods are grounded on obtaining such knowledge - which is often costly, or even impossible. In addition, designing a bias-oriented auxiliary model may not be so intuitive, depending on the type of bias.

\paragraph{Without domain knowledge.} 
Using the fact that malignant biases are learned faster than salient task-relevant features, \citet{lff} train an auxiliary model 
using generalized cross entropy \citep{gce} to emphasize malignant bias and assign more weight to the bias-free samples for the debiased follower. 
\citet{pseudoattr} cluster the training samples 
and reweight the loss of each sample in a batch according to the assigned clusters.
\citet{choo} proposed feature-level data augmentation that disentangles bias features by using a biased auxiliary model obtained by following \citet{lff} and swaps latent bias features. 
\citet{kim2021biaswap} employs task-related features in abundant bias-aligned samples by synthesizing a new image that takes the bias-irrelevant core features from the biased sample and the bias attribute from the bias-free sample.
However, all these methods presume that bias-free samples do exist in sufficient quantities and can be distinguished, which cannot always be guaranteed.

\paragraph{Contrastive learning} is a self-supervised learning method proven to learn representations substantially beneficial to numerous downstream tasks, achieving state-of-the-art performance~\citep{simclr, moco_v2, byol, simsiam, supcon}. \citet{simclr} defines the contrastive prediction task in two steps: a) augmenting two views from the same image by strong data augmentations~\citep{aug_vision_1, aug_vision_2, aug_vision_3} and b) maximizing agreement between the augmented views on the latent space by employing the contrastive loss~\citep{infonce} to capture core features of the image. In the process of maximizing the agreement between the augmented views, the encoder discards the deviating features between the views and learns transformation-invariant representations. For example, using color distortion methods such as \textit{color jitter} and \textit{gray scale} in the augmentation step encourages the encoder to learn color-invariant representations.

\section{Conclusion} \label{sec:conclusion}
In this paper, we propose a novel debiasing method called Contrastive Debiasing via Generative bias-transformation (CDvG). This method is a general debiasing approach that does not require bias labels, bias type information, or even the existence of bias-free samples. Motivated by the observation that not only discriminative models but also image translation models tend to focus on bias, we utilize a translation model as a bias-translator to synthesize novel views with altered bias features. We then employ contrastive learning to pit one bias type against another and obtain bias-invariant representations. CDvG can be integrated with other debiasing techniques in a plug-and-play manner and especially shows good synergy when integrated with models that focus on bias-free samples, such as \citet{lff}. Our extensive experiment on various datasets shows that our method outperforms state-of-the-art, especially when bias-free samples are scarce or even absent.

As a future direction, and a current limitation, we aim to extend our method to datasets where bias-free data are the majority. Recent debiasing approaches, including our method, focus on the highly biased setting, which has yet to be conquered. While these situations may not require debiasing techniques from the start, suggesting a robust method that works in general is an important area of ongoing research.


\section*{Acknowledgements}
This work was supported by Institute of Information \& communications Technology Planning \& Evaluation (IITP) grants (No.2019-0-00075, Artificial Intelligence Graduate School Program (KAIST), No.2022-0-00713, Meta-learning applicable to real-world problems), and National Research Foundation of Korea (NRF) grants (No.2018R1A5A1059921, No.RS-2023-00209060, A Study on Optimization and Network Interpretation Method for Large-Scale Machine Learning) funded by the Korea government (MSIT).

\bibliography{icml_ref}
\bibliographystyle{icml2023}

\newpage
\appendix
\onecolumn
\section{Datasets} \label{appendix:datasets}

\textbf{Colored MNIST} is a biased version of MNIST with colors as biases. 

\textbf{Biased MNIST} is a biased version of MNIST with multiple biases such as digit color, type of background texture, background color, co-occurring letters, and colors of the co-occurring letters.

\textbf{Corrupted CIFAR-10} is an artificially corrupted version of CIFAR-10~\citep{cifar10} to carry biases as proposed in~\citet{robust}. Specifically, the dataset has been corrupted by the following types of method: \{Snow, Frost, Fog, Brightness, Contrast, Spatter, Elastic transform, JPEG, Pixelate, and Saturate\}.

\textbf{Biased Action Recognition~(BAR)} is a real-world action dataset proposed by \citet{lff}. There are six action labels that are biased toward background places.

\textbf{BFFHQ} is proposed by \citet{kim2021biaswap} which is curated from Flickr-Faces-HQ~\citep{ffhq}. It consists of face images where age (young/old) is a task label and gender (male/female) is a bias attribute.

\textbf{ImageNet-9 (IN-9)} is proposed by \citet{rebias} which is a subset of ImageNet~\citep{russakovsky2015imagenet} containing 9 super-classes with texture as biases~\citep{ilyas2019adversarial}.

\textbf{Waterbirds} is proposed by \citet{GroupDPO}, which combines bird photographs from the Caltech-UCSD Birds-200-2011 (CUB) dataset with image backgrounds from the Places dataset. It consists of bird images where a type of bird is a task label and a background (water/land) is a bias attribute.

Colored MNIST, Corrupted CIFAR-10, and BFFHQ can be obtained from \href{https://github.com/kakaoenterprise/Learning-Debiased-Disentangled}{the official Github repository of DisEnt \citep{choo} (https://github.com/kakaoenterprise/Learning-Debiased-Disentangled)}.

Biased MNIST is available in \href{https://github.com/erobic/occam-nets-v1}{the official Github repository of OccamNets~\citep{occamnet} (https://github.com/erobic/occam-nets-v1)}.

BAR is available in \href{https://github.com/alinlab/BAR}{the BAR Github repository (https://github.com/alinlab/BAR)} provided by \citet{lff}. 

ImageNet-9 is available in \href{https://github.com/clovaai/rebias}{the official Github repository of ReBias~\citep{rebias} (https://github.com/clovaai/rebias)}. 

Waterbirds is available in \href{https://github.com/kohpangwei/group_DRO}{the official Github respository of GroupDRO~\citep{GroupDPO} (https://github.com/kohpangwei/group\_DRO)}.


\section{Implementation details} \label{appendix:implementation}
Basically, we follow the same experimental settings in baselines \citep{lff, choo}.  All models are trained on 4 RTX-3090ti GPUs.

\textbf{Training StarGAN.} First, we specify the details of training biased StarGAN. We basically follow the default settings for architectures, optimizers weights for loss terms, and other training configurations in \href{https://github.com/yunjey/stargan}{the Github repository of StarGAN (https://github.com/yunjey/stargan)} across all datasets. For the generator, we use the basic architecture which is composed of total of 3 and 6 blocks for Colored MNIST and the others, respectively. Each block consists of 2 convolutional layers and a skip connection. For the discriminator, we use the architecture which is composed of total 
4 and 5 blocks for \{Colored MNIST
, BFFHQ and IN-9\} and \{Corrupted CIFAR10 and waterbirds\}, respectively. For Colored MNIST, we set the reconstruction weight to 500 and trained for 5000 iterations without random horizontal flipping. 
For 
IN-9 and Waterbirds, we resize them to 224 
and use the original image size for others.

\textbf{Training Configuration.} For the encoder, we use MLP with three hidden layers for Colored MNIST, randomly initialized ResNet-18~\citep{resnet} for Corrupted CIFAR-10 and BFFHQ 
We use batch sizes of 256, 128, and 64 for \{Colored MNIST, and Corrupted CIFAR-10\}, IN-9, and \{BFFHQ and waterbirds\}, respectively. We use learning rates of 0.001 
 for \{Colored MNIST, Corrupted CIFAR-10, BFFHQ, IN-9 and waterbirds\}.
 Also, We use the Adam optimizer with default parameters. We train the model for 
 120, 200, and 500 epochs for 
 IN-9,\{Colored MNIST, BFFHQ and waterbirds\} and Corrupted CIFAR-10, respectively. We use cosine annealing from initial learning rates $lr$ to $lr * 0.1^3$ for learning rate scheduling~\citep{cosine} for all datasets. Note that we do not use random horizontal flipping for ColoredMNIST. Additionally, the original image size of BFFHQ is 128, but we resize them to 224 by following the previous work \citep{choo}.

\textbf{Contrastive learning.} we use Normalized Temperature-scaled Cross Entropy \textit{NT-Xent}~\citep{simclr} with a temperature parameter 0.01. For projection head $H$,  2-layer MLP and a linear layer with the dimensions from the input to the output as [512, 512, 128] and [100, 100] for \{Corrupted CIFAR-10, BFFHQ
, IN-9 and waterbirds\} and Colored MNIST respectively, following \citet{simclr} for the most of the settings.


\textbf{Evaluation.} Note here that, by following \citet{choo}, the performances on BFFHQ whose task is a binary classification, are evaluated only on bias-free test samples that consist of young male and old female samples.

\section{Learning without Bias-free Samples}\label{sec:supp_withoutbiasfreesamples}

In Table \ref{table:extreme}, we empirically demonstrate that previous methods fail to effectively debias the
model (showing low classification scores, almost comparable to Vanilla) when the bias-free samples are absent, i.e., when the biases are more malignant.
\citet{lff} and \citet{choo} show surprisingly degraded performances due to the failure of reweighting scheme. Also, \citet{kim2021biaswap} and \citet{youngyu} break down because they require bias-free samples necessarily to construct pairs of the bias-aligned and bias-free samples in the same class.

\begin{table}[h]
\vspace{-0.1in}
\caption{\small Comparison of debiasing methods on three biased datasets devoid of bias-free samples~(ratio 0\%). We report averaged accuracy on the last epoch and the standard deviation over 3 runs.}
\setlength{\columnsep}{1pt}
\begin{center}
\begin{scriptsize}
\setlength{\tabcolsep}{0.7pt} 
\renewcommand{\arraystretch}{1.2} 
\resizebox{0.6\linewidth}{!}{
\begin{tabular}{@{\extracolsep{3pt}}lccc@{}}
\toprule
\textbf{Method} & ColoredMNIST & CorruptedCIFAR10  & BFFHQ \\ 
\cline{2-4}
\textbf{(ratio (\%))} & 0 & 0 & 0 \\
\cline{1-4}
                     Vanilla & 12.53\tiny{$\pm 0.92$} & 16.05\tiny{$\pm 0.13$}& 37.93\tiny{$\pm 0.96$}\\
                     \cdashline{1-4}
                     LfF \citep{lff} 
                     & 13.16\tiny{$\pm 1.87$} 
                     & 15.88\tiny{$\pm 0.45 $}
                     & 39.67\tiny{$\pm 1.00$} \\
                     DisEnt \citep{choo}
                     & 11.65\tiny{$\pm 0.61$}
                     & 18.76\tiny{$\pm 0.88$}
                     & 38.13\tiny{$\pm 2.13$} \\
                     ReBias \citep{rebias}
                     & 12.72\tiny{$\pm 0.07$}
                     & 14.40\tiny{$\pm 0.49$}
                     & 38.73\tiny{$\pm 0.38$} \\
                     SoftCon \citep{youngyu}
                     & 34.06\tiny{$\pm 0.41$}
                     & 25.90\tiny{$\pm 0.22$}
                     & 38.65\tiny{$\pm 0.38$} \\

                     \cline{1-4}
                     $\text{\textbf{CDvG}}$ & \textbf{95.97}\tiny{$\pm 0.18$} & \textbf{31.24}\tiny{$\pm 0.31$} & \textbf{42.80}\tiny{$\pm 0.33$}   \\
\bottomrule
\end{tabular}
}
\end{scriptsize}
\end{center}
\label{table:extreme}
\end{table}

\begin{figure}[b]
\centering
 \includegraphics[width=0.36\linewidth]{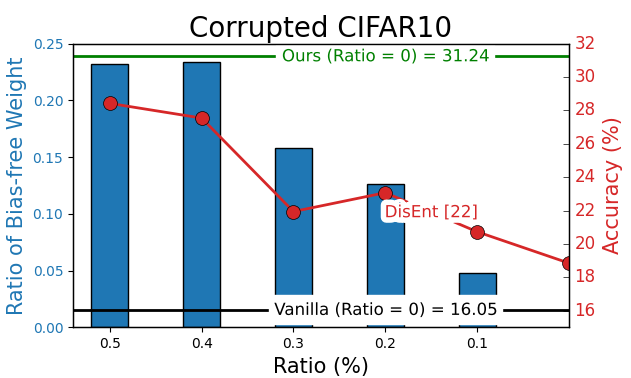}
 \caption{\small Changes in the ratio of bias-free weight~(blue bar) and accuracy~(red line) of DisEnt~\citep{choo} as the ratio of bias-free samples~(x-axis) decreases. The green and black horizontal lines are the accuracy of ours (GDvG) and Vanilla, respectively, when the bias-free samples are absent.}
\label{fig:skewed_choo}
\end{figure}

Furthermore, to evaluate whether the reweighting scheme assigns sufficiently high weights to bias-free samples when bias-free samples are scarce, we further analyze DisEnt~\citep{choo}, which is the state-of-the-art, on Corrupted CIFAR10. In Figure \ref{fig:skewed_choo}, we measure the ratio of bias-free weight (blue bar), which implies how much the bias-free samples are focused, as the ratio of the sum of weights for bias-free samples to the total sum of weights for all samples $\frac{\sum_{i\in F} w_i}{\sum_{i=1}^N w_i}$ where $w_i$ is the assigned weight of $i$-th sample, $F$ is the set of indices of bias-free samples and $N$ is the total number of training samples. As the bias-free samples become scarce, the ratio of bias-free weight calculated by DisEnt decreases, which fails to debias the model, resulting in decreased accuracy (red line).

On the other hand, CDvG shows better performance even with 0\% bias-free samples compared to the DisEnt with 0.5\% bias-free samples. This is because CDvG does not necessarily require bias-free samples for debiasing, unlike the baselines~\citep{rebias, youngyu, lff, choo, kim2021biaswap}.

\section{Bias-transformed images by CycleGAN}

As an example that other translation models also can be used in our framework, we show bias-transformed images generated by CycleGAN~\citep{cyclegan}, which translates between two domains trained on BAR dataset.

The leftmost column of the figure contains the original images and each column is the transformed images of each target domain. The resulting images show the translated background bias attributes.

\begin{figure}[h]
\centering
\includegraphics[width=0.85\linewidth]{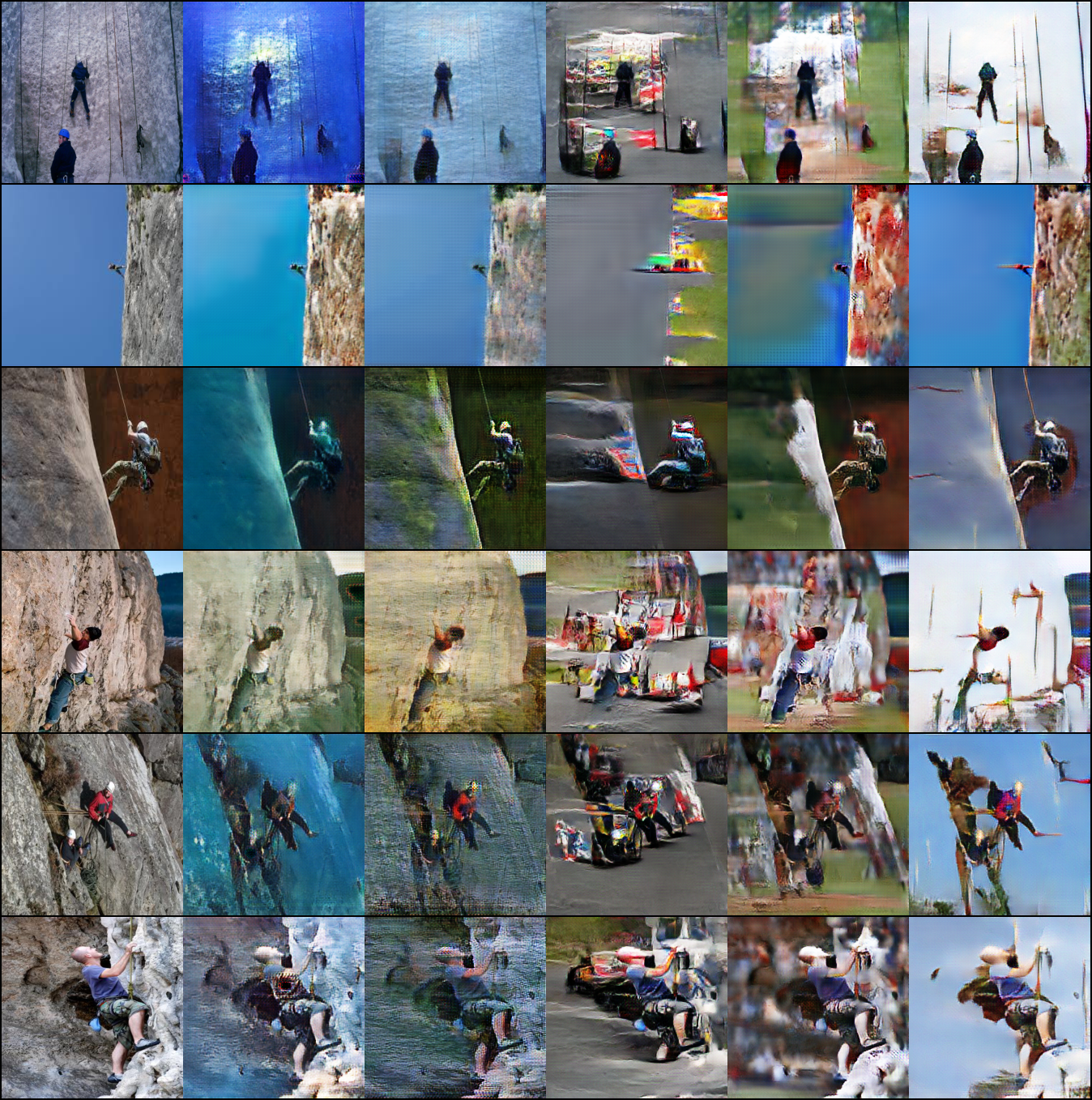}
\caption{\small Bias-transformed images by CycleGAN on BAR dataset.}
\label{fig:appendix_bias-transformed images}   
\vspace{-0.1in}
\end{figure}

\clearpage
\section{StarGAN V2}
\label{appendix:starganv2}
We conducted an experiment in which the multi-task discriminator of \citet{starganv2} was replaced with a domain discriminator. In Table~\ref{table:starganv2}, we empirically demonstrate that recent translation models are also inherently bias-susceptible to bias when using a domain classifier. In conclusion, the use of recent translation models can improve the debiasing performance by improving the quality of translated images, but there is a trade-off between cost and performance due to the large size of the network (6 times larger than StarGAN). 

\renewcommand{\arraystretch}{1.3}
\begin{table}[h]
\caption{Corrupted CIFAR-10 with \citet{starganv2}}
\vspace{-0.1in}
\begin{center}
\resizebox{0.5\linewidth}{!}{
\begin{tabular}{ccccc}
\toprule
{Dataset} &  \thead{Ratio\\(\%)} & Vanilla & \thead{CDvG\\w/ StarGAN} & \thead{CDvG\\w/ StarGANv2} \\
\cline{1-5}
\rule{0pt}{2ex}
\multirow{2}{*}{\thead{Corrupted\\CIFAR-10}} & 0.0 & 20.87 & 31.50 & \textbf{32.30} \\
& 5.0 & 41.12 & 42.75 & \textbf{47.80} \\
\bottomrule
\end{tabular}
}
\end{center}
\label{table:starganv2}
\vspace{-0.2in}
\end{table}
\label{appendix:starganv2} 

\newpage

\newpage


\end{document}